%% file: main.tex
\documentclass[11pt,letterpaper]{article}
\input{preamble}

\title{RoboGrade: Measuring Robotic Execution Quality Beyond Binary Success}
\author{%
\begin{tabular}{@{}c@{\quad}c@{\quad}c@{\quad}c@{}}
Zaruhi Navasardyan\thanks{Equal contribution.} &
Tatul Danielyan\footnotemark[1] &
Mark Hamazaspyan\footnotemark[1] &
Hrant Davtyan \\[0.35em]
\multicolumn{4}{c}{\small\texttt{\{zaruhi, tatul, mark, hrant\}@metric.am}}
\end{tabular}%
}
\date{}

\begin{document}
\maketitle
\input{sec/0_abstract}

\section{Introduction}
\label{sec:introduction}

Robotic manipulation policies are commonly evaluated using task success rate~\cite{liu2023libero,brohan2023rt1}, which overlooks important differences in execution quality. Successful executions may differ in harmful events, their severity, and retries, while failures vary in progress and final-state recoverability. These distinctions matter for diagnosing failures, comparing policies, and selecting training demonstrations, yet cannot be captured by binary success alone.

\begin{figure*}[t]
    \centering
    \includegraphics[width=\textwidth]{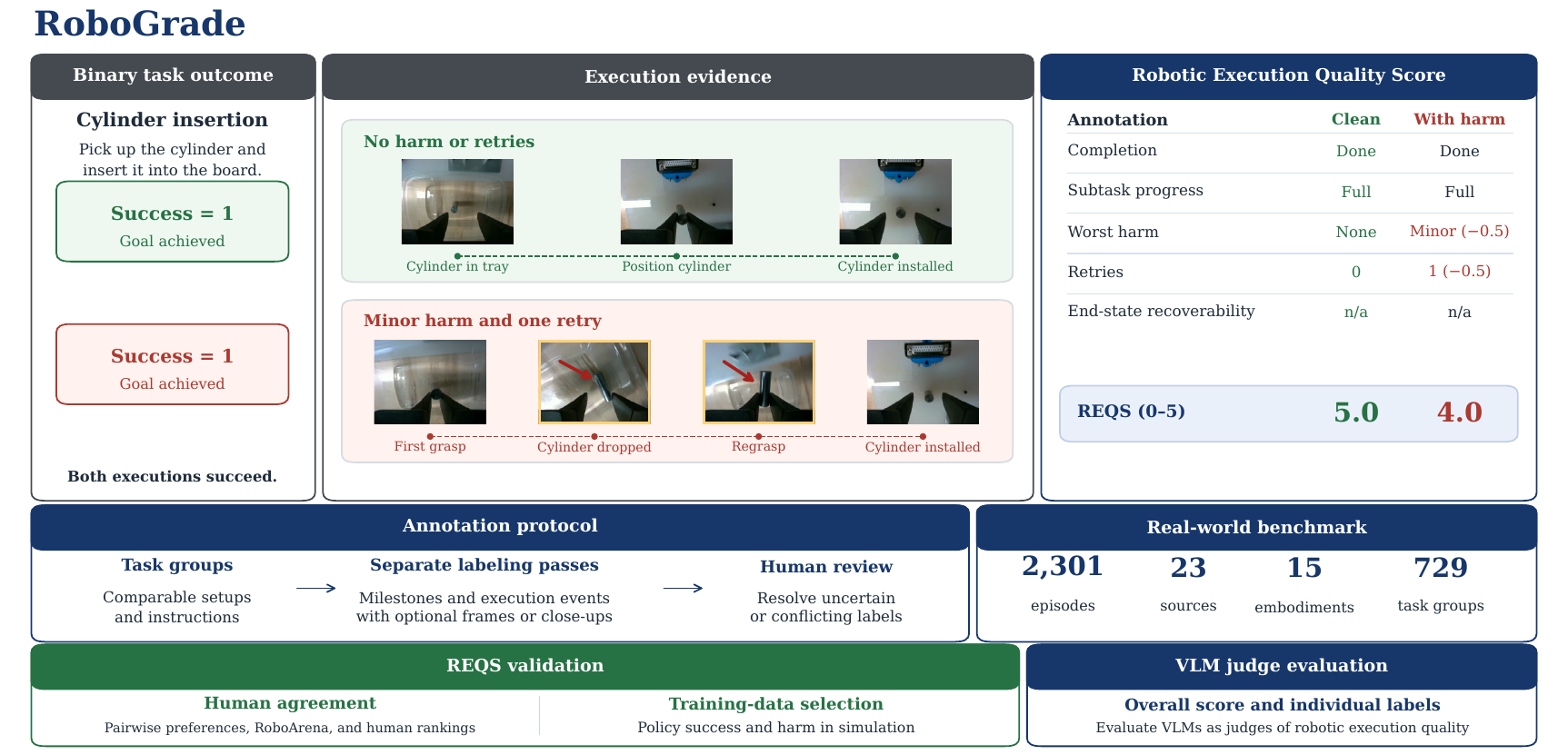}
    \caption{\textbf{RoboGrade measures execution quality beyond binary success.} Both executions install the cylinder, but the second involves a drop and retry, resulting in a lower REQS (4.0 vs.\ 5.0). A shared protocol annotates completion, progress, harm, retries, and recoverability across real-world manipulation episodes. We validate REQS through human judgments and demonstration selection, and benchmark VLMs as execution-quality judges.}
    \label{fig:robograde-teaser}
\end{figure*}

Prior work evaluates robotic execution through human preferences, progress, and fine-grained metrics. RoboArena uses human pairwise comparisons~\cite{atreya2025roboarena}; RoboReward and PRM-as-a-Judge assess task completion and progress from video~\cite{lee2026roboreward,ji2026prmjudge}. RoboEval measures efficiency, safety, stability, and coordination in instrumented simulation~\cite{wang2025roboeval}, while Eval-Actions combines video and robot kinematics to assess execution quality~\cite{liu2026evalactions}. However, existing approaches differ in objectives and input requirements, and a unified, task-grounded assessment of harm severity, retries, and final-state recoverability across heterogeneous real-world recordings remains underexplored.

We introduce RoboGrade, a benchmark and evaluation framework for robotic execution quality (Figure~\ref{fig:robograde-teaser}). It comprises approximately 2,300 real-world manipulation episodes from 23 sources, spanning 15 robotic embodiments and 729 task groups. Using task instructions and video, a shared annotation protocol captures task completion, subtask progress, harmful events and their severity, retries, and final-state recoverability, including event timestamps. We further introduce the Robotic Execution Quality Score (REQS), an interpretable 0--5 metric combining completion and progress with penalties for harm, retries, and unrecoverable failures. Unlike state-dependent approaches, the framework supports evaluation across heterogeneous real-world recordings without requiring robot states or kinematics.

We validate REQS against human preferences and through downstream demonstration selection. REQS achieves approximately 80\% agreement with human pairwise judgments, approaching inter-annotator agreement, and we compare its rankings with those of individual components and learned reward models. In simulated imitation learning, REQS-based demonstration selection improves data efficiency and reduces harmful executions, particularly at smaller training-data scales.

We also benchmark eleven open-weight and proprietary VLMs as execution-quality judges on RoboGrade samples. The best model achieves 58.0\% exact REQS agreement, while balanced accuracy for harm severity and retry counts reaches only 57.1\% and 58.4\%, respectively. These findings reveal a gap between recognizing task outcomes and assessing the execution details underlying overall quality.

Our main contributions are:
\begin{itemize}
    \item \textbf{RoboGrade benchmark.} A unified benchmark of approximately 2,300 real-world episodes from 23 sources, 15 embodiments, and 729 task groups, with multidimensional, video-based execution annotations.

    \item \textbf{Robotic Execution Quality Score (REQS).} An interpretable, task-grounded metric capturing completion, progress, harm, retries, and recoverability, validated against human preferences and through downstream demonstration selection.

    \item \textbf{VLM-based execution quality assessment.} An evaluation of eleven VLMs as judges of overall execution quality and its individual dimensions, revealing limitations beyond task-completion recognition.
\end{itemize}

We release the benchmark, annotations, scoring rule, and evaluation framework.

\section{Related Work}
\label{sec:related-work}

Prior work assesses robotic execution through human preferences and explicit measurements of task performance. RoboArena aggregates human preferences from double-blind policy comparisons on real robots, capturing differences that may not be reflected in binary task outcomes~\cite{atreya2025roboarena}. Simulation enables more direct measurement of execution properties. SIMPLER provides simulated environments for evaluating policies with demonstrated correlations to real-world performance~\cite{li2025simpler}, while RoboEval measures bimanual task progress, coordination, trajectory efficiency, and collision-related behavior using robot and environment state~\cite{wang2025roboeval}. Platforms such as ManiSkill3 also provide task-specific rewards and success conditions computed from simulator state~\cite{tao2025maniskill3}. These approaches establish useful evaluation criteria, but preference rankings do not by themselves specify how individual execution properties contribute to quality, and state-dependent measurements require information unavailable in many recorded videos. REQS provides an explicit scoring rule over components annotated from task instructions and video, enabling consistent assessment across heterogeneous real-world recordings.

Video-based reward models provide a means of assessing execution without access to simulator state. RoboReward introduces a dataset, benchmark, and reward models for predicting completion and partial progress, using counterfactual task relabeling and temporal clipping to augment successful demonstrations with negative and partial-progress examples~\cite{lee2026roboreward}. Robometer combines frame-level progress supervision with trajectory-level preferences to learn from successful and failed executions~\cite{liang2026robometer}. NVIDIA's Flora integrates temporal segmentation, action captioning, and reward prediction; Flora-Rewarder supports episode-level outcome rewards, dense action-quality scores, and scoring under user-defined rubrics~\cite{leng2026flora}. PRM-as-a-Judge derives efficiency, regression, and stagnation measures from dense progress curves~\cite{ji2026prmjudge}, while its extension adds assessments of near-success failures, progress regained after setbacks, and successful execution quality~\cite{liu2026prmjudge15}. Eval-Actions combines completion, motion smoothness, visible collision-related events, and efficiency through expert grading and ranking-calibrated quality labels~\cite{liu2026evalactions}. REQS builds on this broader view of execution quality by explicitly combining completion and subtask progress with penalties for harm severity, retry counts, and unrecoverable failures. Its recoverability component concerns whether completion remains feasible from the final state, which is distinct from measuring recovery observed earlier in the trajectory.

Prior work also provides benchmarks for evaluating the reliability of automated execution judges. FailBench evaluates whether VLMs can distinguish successful from failed manipulation attempts across diverse real and simulated sources, and reports substantially lower accuracy on contact-intensive tasks~\cite{navasardyan2026failbench}. RoboReward evaluates reward prediction across real-world tasks and embodiments, and RoboPulse and RoboPulse++ test recognition of progress and regression through observation pairs and temporal intervals~\cite{lee2026roboreward,ji2026prmjudge,liu2026prmjudge15}. Flora-Rewarder is evaluated on outcome scoring and trajectory ranking using RoboReward and Robometer~\cite{leng2026flora}. AutoEval, the reference evaluator introduced with Eval-Actions, predicts quality scores and diagnostic explanations using RGB observations and robot kinematic summaries~\cite{liu2026evalactions}. Our evaluation centers on REQS: we assess its alignment with human pairwise preferences and its utility for demonstration selection, then examine whether VLMs can predict the score and its constituent dimensions. RoboGrade supports this evaluation with a shared annotation protocol across heterogeneous real-world sources, including component labels and event timestamps. Unlike simulation-based evaluation frameworks that rely on privileged environment states, RoboGrade supports multidimensional execution quality assessment directly from task instructions and real-world videos. While existing video-based approaches primarily emphasize task completion, progress, or trajectory preferences, RoboGrade explicitly incorporates harmful events and their severity, retries, and final-state recoverability into a unified scoring rule. This enables consistent, interpretable evaluation across heterogeneous real-world datasets without requiring simulator states or robot kinematics.

\section{RoboGrade Benchmark}
\label{sec:benchmark}
\subsection{Benchmark Description and Statistics}
RoboGrade brings together real-world manipulation videos under a common rubric for assessing both task outcome and execution quality. The benchmark covers approximately 2,300 episodes from 23 sources, spanning 729 task groups and 15 robotic embodiments, including single-arm, bimanual, and humanoid platforms. The 23 sources include HRDexDB~\cite{hrdexdb2026}, RoboVAD~\cite{robovad2026}, AgiBot World 2026~\cite{agibotworld2026}, ArmnetBench~\cite{armnetbench2026}, RoboChallenge~\cite{robochallenge2025}, ARMBench~\cite{armbench2023}, RH20T~\cite{rh20t2024}, REBOOT~\cite{reboot2026}, REASSEMBLE~\cite{reassemble2025}, BotFails~\cite{botfails2026}, selected clips from research project pages (Other), FailTime-Long~\cite{failtimedata}, Robometer~\cite{liang2026robometer}, RoboReward~\cite{lee2026roboreward}, ViFailback~\cite{vifailback2025}, TOTO~\cite{toto2023}, RoboArena~\cite{atreya2025roboarena}, HapticWAM~\cite{hapticwam2026}, PhAIL~\cite{phail2026}, RoboFAC~\cite{robofac2025}, Handover Failure Detection (HFD)~\cite{hfd2024}, RoboFail (REFLECT)~\cite{reflect2023}, and selected clips from humanoid competition recordings. Placement, transfer, and sorting appear in 48.5\% of episodes, and grasping and lifting in 30.6\%, based on action terms in the instructions. An episode can contribute to multiple task families. Appendix~E lists the source datasets and the number of episodes drawn from each.

\begin{figure*}[t]
    \centering
    \includegraphics[width=\textwidth]{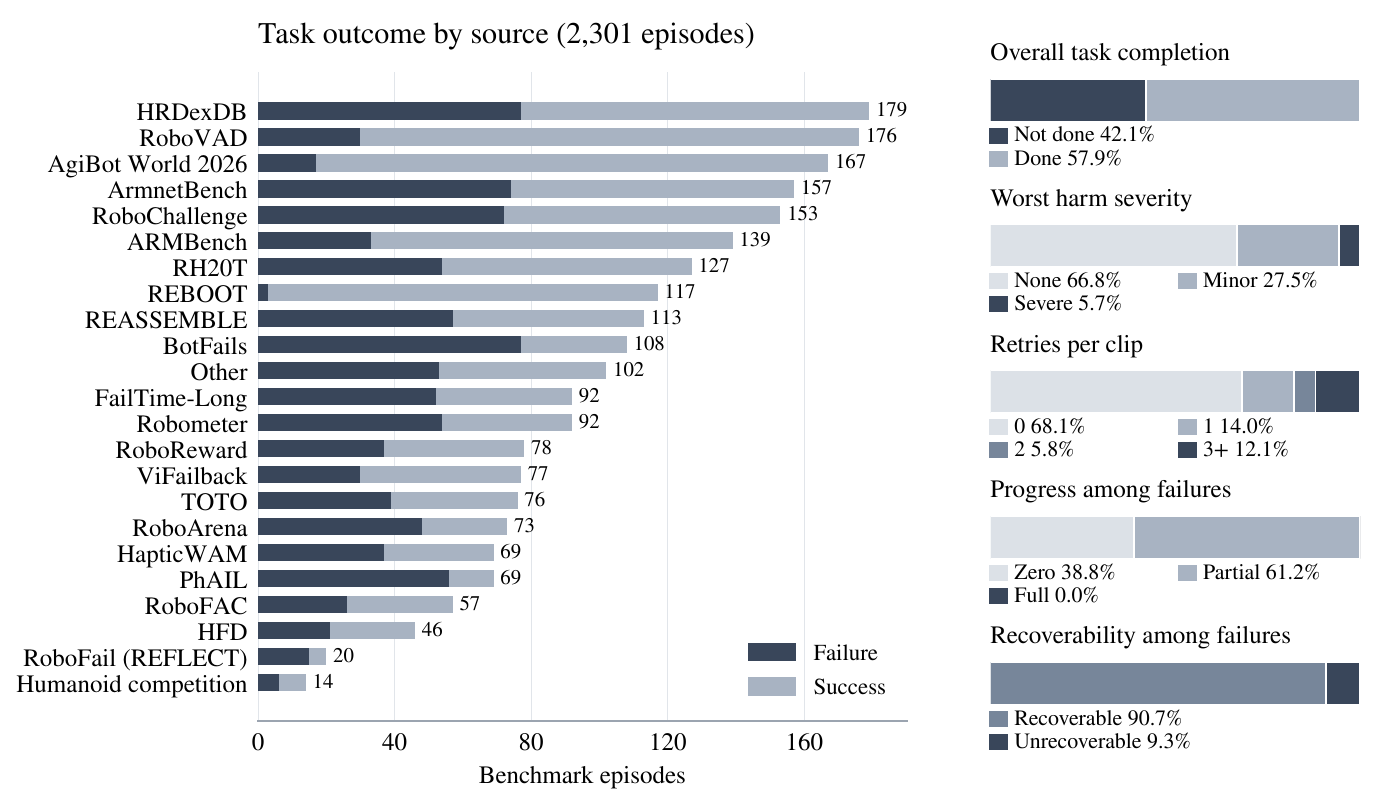}
    \caption{Composition of RoboGrade (2,301 episodes across 15 embodiments). The bars on the left show successful and failed executions by source. Task completion, worst harm severity, and retries are summarized over all benchmark episodes. Progress and recoverability are summarized over the 968 failed executions.}
    \label{fig:benchmark-overview}
\end{figure*}

Figure~\ref{fig:benchmark-overview} shows that the success rate varies across sources and reaches 58\% overall. Across the benchmark, 33.2\% of executions contain a harmful event and 31.9\% include at least one retry. These events also occur in successful executions: 38.9\% of successes involve harm, a retry, or both. Among failed executions, 61.2\% achieve partial progress and 90.7\% end in a state judged recoverable. A success label alone therefore leaves substantial variation within both outcome groups.

\subsection{Annotation Dimensions and Pipeline}
The source datasets differ in both annotation coverage and label definitions. Some provide only an episode outcome or anomaly flag, while others record subtask completion, error timing or harmful events. Success criteria also vary across sources. Depending on the dataset, an execution counts as successful if it reaches the task goal, contains no annotated anomaly, or meets a threshold on a task-specific score. We apply a shared annotation rubric to make these judgments comparable across sources. Supplementary Appendix~A gives examples of these differences and defines each RoboGrade label.

\paragraph{Task completion and progress.}
We define task completion by whether the execution satisfies the goal in the task instruction. To measure partial progress consistently, we define a shared subtask checklist of observable milestones for executions of the same task recorded in comparable settings within a source. Progress is the fraction of applicable milestones achieved. Lifting counts as soon as the object is lifted. A placement milestone requires the robot to release the object and the object to remain in its intended location at the end of the execution. Completion and subtask progress are judged separately, so an inconsistency between them can flag an execution for review.

\paragraph{Execution quality.}
The rubric distinguishes harm, retries, and recoverability because each captures a different aspect of how the task unfolds. Harm records unintended consequences such as collisions, spills, or damage, with severity reflecting their effects. A retry is a renewed attempt at a step. Both event types are timestamped when their onset is visible. For incomplete executions, recoverability describes whether the robot could still finish from the final state without human intervention. The rubric also records observed intervention separately.

GPT-6.1 Sol applied the rubric using the task instruction and available camera views, requesting closer views or additional frames when the evidence was ambiguous. Across benchmark episodes, the execution annotations took one, two, or three turns in 35.0\%, 42.1\%, and 22.9\% of cases, respectively. Close-ups were requested for 30.1\% of episodes and additional frames for 59.8\%. Supplementary Appendix~A details how recording setups are grouped and how follow-up evidence is requested.

\subsection{Annotation Quality Control}
We flagged executions for human review in three cases:
\begin{itemize}
    \item The model marked a label as uncertain.
    \item An annotation disagreed with a corresponding source label.
    \item Task completion and the subtask labels disagreed, either because all applicable milestones were marked complete for a failed execution or because a required milestone was marked incomplete for a successful execution.
\end{itemize}
Reviewers removed the flagged execution, revised its labels, or accepted the original annotations. We also audited 400 (25\%) unflagged labels to assess agreement on cases that passed these checks. Model--reviewer agreement on this audit was 90\% (for 90\% of episodes no labels were changed). Overall, 50\% of the annotation were human reviewed, all the rest are unflagged samples. Appendix~A reports agreement by annotation dimension for the human-reviewed benchmark episodes.

\section{Robotic Execution Quality Score (REQS)}
\label{sec:reqs}
\subsection{Metric Definition and Formulation}

Executions with identical task outcomes may differ substantially in their quality. To capture these distinctions, we introduce the \textbf{Robotic Execution Quality Score (REQS)}, a unified and interpretable metric for assessing robotic execution quality. REQS provides a common scoring rule for comparing executions while retaining the contribution of each dimension, making score differences interpretable. The relative importance of individual dimensions can also be adjusted to reflect application-specific priorities.

REQS assigns execution quality on a 0--5 scale. Task completion and intermediate progress determine the base score, while harmful events, retries, and an unrecoverable final state introduce penalties.

Let $d$ indicate task completion, and let $p\in[0,1]$ be the fraction of applicable milestones achieved for an incomplete execution. A completed task starts at 5 points, while an incomplete task starts at $4p$. Let $h$ denote the worst harm severity, $r$ the number of retries, and $u$ an indicator that the execution ended in an unrecoverable state. We set $u=0$ for completed tasks. The score before applying bounds is
\begin{equation}
s = B(d,p) - H(h) - R(r) - u,
\end{equation}
where $B(1,p)=5$ and $B(0,p)=4p$. The harm penalty $H$ is 0 for no harm, 0.5 for minor harm, and 2 for severe harm. The retry penalty $R$ is 0 for no retries, 0.5 for one or two retries, and 1 for three or more. We calibrated the scoring weights through iterative human feedback on candidate scores for a sample of executions.

We bound the score to $[0,5]$ and set a minimum of 1 for executions that complete the task or achieve any subtask progress. Appendix~B of the supplementary material provides worked examples.

\subsection{Human Agreement}
\label{sec:human-pref}

We assess REQS against human judgments in three ways. Independent pairwise ratings establish how REQS agreement compares with agreement between human raters. RoboArena preferences and written explanations help identify where REQS and evaluators differ. A rank-based comparison tests how closely REQS and alternative metrics, including learned rewards, reproduce human rankings.

Two annotators independently compared 200 pairs of robot executions. Within each pair, both executions addressed the same task in the same setup. The pairs spanned small and large differences in REQS. Agreement between REQS and the human preferences averaged 80\% across the two annotators, compared with 88\% agreement between annotators. Manual review of the disagreements indicated that annotators sometimes overlooked brief retries or harmful events when comparing the two videos. This suggests that annotating these events separately can capture details that are easy to miss in an overall comparison. Reviewers also cited execution speed in their explanations. REQS captured some of these preferences when the faster execution required fewer retries. Speed can also differ between executions with the same retry count, leaving some speed-based preferences unexplained by the score.

To examine where REQS agrees with human preferences across different task outcomes, we also compared its rankings with RoboArena judgments~\cite{atreya2025roboarena}. When one rollout succeeded and the other failed, agreement was 91\%. The disagreements reflected differences in the definition of success. For example, an evaluator could count lifting an object as success even if the robot subsequently dropped it, whereas REQS assesses whether the required goal state holds at the end. When both rollouts failed, REQS reproduced 85\% of evaluator ties and 45\% of expressed preferences. Evaluators sometimes credited early actions, such as reaching for the correct object, before either rollout achieved a subtask.

Among successful rollouts that RoboArena evaluators judged equivalent, REQS reproduced 35\% of ties, compared with 83\% for the progress-only baseline. Retries and minor harm accounted for most of the disagreements, suggesting that evaluators often gave these events little weight when judging two successful executions as equivalent. When evaluators expressed a preference despite assigning both successful rollouts the same progress score, the penalties increased agreement from 20\% to 70\%.

We further examined how individual REQS components relate to human rankings within a task. A human reviewer ranked 10--15 executions for each of 3 tasks. For each task, we computed Spearman's rank correlation between the human ordering and rankings derived from REQS, its individual dimensions, and two learned reward models, Robometer-4B and RoboReward-8B~\cite{liang2026robometer,lee2026roboreward}. The reviewer consistently ranked successful executions above failed ones. To examine what explained preferences within a shared outcome, we analyzed successful and failed executions separately and evaluated subtask progress as a ranking criterion only among failures. Subtask progress was the strongest individual correlate of human rankings among failed executions. Among successful executions, fewer retries aligned most closely with the human ranking. These patterns support a composite score that retains progress when the goal is unmet and accounts for retries and harm when the goal is achieved.

\begin{table}[t]
    \centering
    \small
    \caption{Spearman's rank correlation $\rho$ with human rankings,
    averaged over three tasks (10--15 executions each) for which the
    ranking score varies. Harm and retries are ordered from least to
    most. Subtask progress ranks successes first in the full set and
    distinguishes executions within failures.}
    \label{tab:human-ranking}
    \setlength{\tabcolsep}{4pt}
    \begin{tabular}{@{}lrrr@{}}
        \toprule
        Ranking score & All & Success & Failure \\
        \midrule
        REQS & \textbf{0.86} & 0.50 & 0.34 \\
        \quad Subtask progress & -- & -- & \textbf{0.45} \\
        \quad Harm severity & 0.16 & 0.27 & $-0.17$ \\
        \quad Retry count & 0.19 & \textbf{0.78} & 0.40 \\
        \midrule
        Robometer-4B~\cite{liang2026robometer} & 0.73 & 0.22 & 0.40 \\
        RoboReward-8B~\cite{lee2026roboreward} & 0.41 & 0.22 & $-0.20$ \\
        \bottomrule
    \end{tabular}
\end{table}

\subsection{Downstream Validation}
\label{sec:reqs-data-selection}
\begin{table*}[t]
    \centering
    \small
    \caption{Test results of BC-ResNet for each training-data selection strategy. Each row is one training-set size, given as the number of time steps (frames).}
    \label{tab:reqs-data-selection}
    \setlength{\tabcolsep}{2.5pt}
    \begin{tabular}{@{}l rrr rrr@{}}
        \toprule
        Time steps & \multicolumn{3}{c}{Success (\%) $\uparrow$} & \multicolumn{3}{c}{Harm (\%) $\downarrow$} \\
        \cmidrule(lr){2-4}\cmidrule(l){5-7}
        (frames) & Random & Success only & Top REQS & Random & Success only & Top REQS \\
        \midrule
        12k & 83.2 & 91.4 & \textbf{95.9} & 47.2 & 40.1 & \textbf{31.9} \\
        24k & 91.4 & 95.4 & \textbf{97.4} & 39.8 & 37.2 & \textbf{31.6} \\
        47k & 99.0 & 98.2 & \textbf{99.2} & 30.7 & 32.5 & \textbf{30.4} \\
        \bottomrule
    \end{tabular}
\end{table*}

We hypothesize that selecting training demonstrations by REQS yields policies with higher task success rates and lower harm rates than selecting demonstrations by success alone, especially when many successful demonstrations contain errors or harmful events.

We collected GR00T N1.7~\cite{nvidia2025gr00t} rollouts on LIBERO-Object~\cite{liu2023libero}, where the robot places a named object in a basket. During the rollouts, we added brief perturbations (premature gripper opening, pushes, sideways shifts, or shaking) to produce episodes with errors and recovery. The resulting pool contains 1,926 rollouts across ten object variants, of which 79\% succeed. We compute REQS for every rollout from simulator state. Harm covers collisions of the robot or the held object with the table or other objects, drops, and objects knocked over or moved. It is detected from contact forces and object poses.

We compare three selection strategies: random sampling, sampling only successful episodes, and selecting the episodes with the highest REQS. In practice, every top-REQS set contains only successful episodes with REQS of at least 4.5, and about 70\% of them score 5. Success-only sets include successes with REQS as low as 3.5, and 57--58\% of their episodes contain harm or retries.

We match the three training sets by the number of time steps (frames), not by the number of episodes, because failed episodes are longer than successful ones. With equal episode counts, the random set would contain more time steps.

We train a behavior-cloning policy, BC-ResNet, which imitates the actions in its training episodes. It uses two ResNet-18 image encoders~\cite{he2016resnet} (front and wrist camera) initialized from ImageNet and predicts the next eight actions. We use batch size 128 and AdamW with learning rate $3\times10^{-4}$ and cosine decay. At each training-set size, the three policies are trained for the same number of updates.

We test each trained policy on 200 held-out scenes, 20 per object, with the objects at new random positions. For each policy, we report the success rate and the harm rate (the share of test rollouts with at least one harmful event). Both are computed from simulator state, in the same way as for the training data.

\begin{figure}[t]
    \centering
    \includegraphics[width=0.65\textwidth]{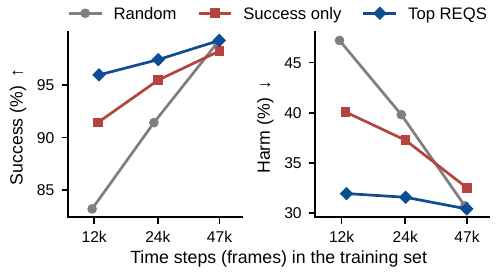}
    \caption{BC-ResNet test success and harm by the number of time steps (frames) in the training set, averaged over training runs.}
    \label{fig:data-amount}
\end{figure}

Compared with success-only selection, top-REQS selection reduced harm at every training-set size (Table~\ref{tab:reqs-data-selection}). The reduction was largest for the two smaller training sets, at 8.2 and 5.6 percentage points. Success was also higher at every size, by 1.0 to 4.5 points.

The gains were largest with the least data (Figure~\ref{fig:data-amount}). Top REQS trained on about 12k time steps matched the success of success-only selection trained on about 24k (95.9\% vs.\ 95.4\%) and caused less harm (31.9\% vs.\ 37.2\%). The main effect of selecting by REQS is lower harm, with a smaller gain in success. This is consistent with how the training sets differ: success-only sets keep successful episodes that contain harm or retries, while top-REQS sets keep mostly clean ones.
Overall, REQS-based selection is particularly effective in data-limited settings, outperforming success-only and random selection even with fewer training samples. At larger data scales, random selection approaches REQS performance, potentially reflecting the increasing importance of demonstration diversity as data availability grows.

\section{VLM Evaluation and Results}

\subsection{Experimental Setup}
The evaluation covers open-weight and proprietary models spanning different sizes and visual and embodied reasoning capabilities: Qwen, Gemma, Cosmos-Reason2-32B, Nemotron 3.5 Super VL, GLM 5.3 Flash, Gemini 3.8 Flash, Claude Opus 5.5, and TypeSafe Jev 1.14~\cite{qwen2026qwen3827b,qwen2026qwen359b,qwen2026qwen354b,gemmateam2026gemma4,nvidia2026cosmosreason2,nvidia2026nemotron35,zai2026glm53flash,google2026gemini38flash,anthropic2026opus55,typesafe2026jev114}. OpenAI models are excluded because GPT-6.1 Sol contributed to the reference annotations, reducing the risk of inflated agreement from shared labeling tendencies~\cite{openai2026gpt61sol}.

Each judge receives the task instruction, subtask checklist, and all available camera views, and predicts the annotation dimensions used to compute REQS. All models undergo one evaluation pass on the same 2,301 clips using a shared rubric without model-specific prompt tuning. Video sampling follows model defaults or documented recommendations; models without native video input receive 32 timestamped frames per view, matching Gemma's frame budget. For decision models, structured-question probabilities are converted to rubric labels using fixed rules. Full prompts, provider configurations, and inference settings are provided in the supplementary material.

Balanced accuracy is reported for completion, subtask achievement, worst harm severity, retry count bins, and recoverability, with recoverability evaluated only on reference failures. Predicted and reference events of the same kind are matched one-to-one within three seconds or one frame interval, whichever is larger. Unmatched predictions make the corresponding clip-level harm or retry label incorrect. A clip receives a predicted REQS only if its predicted harm and retry answers pass the localization check. 

REQS is evaluated using mean absolute error (MAE) and exact agreement. Clip-level resampling provides 95\% confidence intervals.

\begin{table*}[t]
    \centering
    \small
    \caption{Judge evaluation on 2,301 RoboGrade clips. The five annotation columns report balanced accuracy in percent. Exact reports REQS exact agreement in percent. MAE is in REQS points. Parameter counts are totals. Nemotron activates 12B of its 120B parameters, and GLM 5.3 Flash activates 18B of its 320B parameters.}
    \label{tab:vlm-results}
    \setlength{\tabcolsep}{2.5pt}
    \begin{tabular}{llrrrrrrr}
        \toprule
        & & \multicolumn{5}{c}{Annotation dimensions} & \multicolumn{2}{c}{REQS} \\
        \cmidrule(lr){3-7}\cmidrule(lr){8-9}
        Model & Params. & Done & Subtasks & Harm & Retries & Recovery & Exact & MAE \\
        \midrule
        Claude Opus 5.5 & -- & 89.7 & 85.8 & 57.1 & 58.4 & 65.3 & 58.0 & 0.486 \\
        Gemini 3.8 Flash & -- & 86.8 & 85.0 & 54.7 & 50.8 & 64.5 & 54.8 & 0.717 \\
        Qwen3.8-27B & 27B & 84.2 & 81.6 & 51.9 & 52.8 & 59.9 & 51.8 & 0.706 \\
        TypeSafe Jev 1.14 & -- & 80.0 & 75.7 & 42.9 & 37.9 & 50.6 & 31.3 & 0.816 \\
        GLM 5.3 Flash & 320B & 76.9 & 76.8 & 46.5 & 44.4 & 46.3 & 40.1 & 0.913 \\
        Qwen3.5-9B & 9B & 77.0 & 76.2 & 43.7 & 35.7 & 46.9 & 41.2 & 0.973 \\
        Gemma 4 31B & 31B & 74.4 & 75.3 & 42.4 & 42.1 & 52.2 & 39.6 & 1.084 \\
        Qwen3.5-4B & 4B & 75.4 & 72.8 & 42.7 & 35.6 & 44.1 & 38.8 & 1.007 \\
        Nemotron 3.5 Super VL & 120B & 65.1 & 62.8 & 38.5 & 32.3 & 24.8 & 36.8 & 1.397 \\
        Cosmos-Reason2-32B & 32B & 59.5 & 57.7 & 38.1 & 33.0 & 17.1 & 38.0 & 1.573 \\
        Gemma 4 12B & 12B & 51.1 & 47.3 & 34.2 & 31.0 & 15.1 & 32.5 & 1.706 \\
        \bottomrule
    \end{tabular}
\end{table*}

\subsection{Results}
Claude Opus 5.5 achieves the lowest MAE (0.486 REQS points) and the highest exact agreement (58.0\%, Table~\ref{tab:vlm-results}). Qwen3.8-27B leads the open-weight models on both metrics. Its MAE is slightly lower than Gemini 3.8 Flash's (0.706 versus 0.717), while Gemini has higher exact agreement (54.8\% versus 51.8\%).

Nemotron 3.5 Super VL, Cosmos-Reason2-32B, and Gemma 4 12B have the highest MAEs, ranging from 1.397 to 1.706 REQS points. After localization, Gemma 4 12B has no score for 10.6\% of clips and TypeSafe Jev 1.14 has no score for 51.3\%. MAE should be interpreted alongside exact agreement, which counts missing predictions as misses. These results compare complete judge configurations, including input sampling and output reliability, rather than isolating model architecture or training effects.

Claude Opus 5.5 achieves the lowest MAE on 17 of the 23 source datasets (Supplementary Sec.~C), suggesting that its aggregate advantage is broadly distributed across sources. Nevertheless, even its error varies substantially, from 0.05 on REBOOT to 1.24 on ARMBench. ARMBench and REASSEMBLE yield MAEs above one point for every evaluated judge, whereas several models achieve low errors on RoboVAD and ViFailback. Strong aggregate performance therefore does not imply uniform reliability across datasets. These comparisons are descriptive: differences may reflect execution difficulty, reference-score distributions, and prediction coverage, since MAE is computed only over numeric outputs.

Task completion and subtask achievement are generally the strongest dimensions, while harm severity and retry counts remain substantially harder (Table~\ref{tab:vlm-results}). Even Claude Opus 5.5, the strongest overall judge, correctly classifies over 93\% of harm-free and zero-retry clips but only 34.1\% of severe-harm clips and clips with three or more retries. Even the strongest judge therefore frequently misclassifies severe harm and repeated retries under our evaluation protocol.

Event localization presents a further limitation: Claude matches fewer than half of the reference harm and retry events within the temporal tolerance. Recoverability also remains challenging, with balanced accuracy reaching at most 65.3\% on reference failures. For Claude, this rises to 86.7\% when restricted to failures correctly identified as incomplete, suggesting that failure recognition is an important obstacle. However, this selected subset may exclude harder cases and differs across models.

Weak failure recognition also accompanies optimistic scoring. Nemotron 3.5 Super VL, Cosmos-Reason2-32B, and Gemma 4 12B frequently misclassify incomplete executions and repeated attempts, while overestimating REQS by 1.14--1.35 points on average over numeric predictions. This pattern is consistent with overpredicting completion and understating retry penalties, although it does not isolate their individual contributions.

Overall, the results suggest that frontier VLMs can assess robotic task outcomes and progress, as reflected in their relatively strong performance on task and subtask completion. However, evaluation through REQS exposes gaps in assessing harm, retries, and recoverability.

\section{Conclusion}
\label{sec:conclusion}

We introduce RoboGrade, a benchmark and evaluation framework for multidimensional robotic execution quality assessment across heterogeneous real-world recordings, without requiring simulator states or robot kinematics. Its unified annotation protocol and the Robotic Execution Quality Score (REQS) extend binary task success by explicitly capturing progress, harmful events and their severity, retries, and final-state recoverability. We validate REQS through independent human preferences and demonstrate its downstream utility for demonstration selection, improving data efficiency and reducing harmful executions.

Our evaluation of VLM-based judges shows that the strongest models estimate reference REQS scores with relatively low mean absolute error, although exact agreement remains limited. These results suggest potential for automated assessment of overall execution quality, even when precise score prediction remains challenging.

We therefore advocate REQS as a practical, standardized framework for assessing robotic execution quality, with VLM-assisted annotation enabling scalable evaluation directly from task instructions and video. Although fine-grained assessments and safety-critical applications still require additional verification, RoboGrade provides an interpretable and actionable alternative to binary success evaluation, supporting more informative policy comparison, failure analysis, and demonstration selection across real-world robotic tasks.

\section{Limitations and Future Work}

The insights gained from this research are subject to a few limitations that simultaneously point
toward compelling directions for future study.

\textbf{Visual evidence and annotation granularity.}
RoboGrade relies on task instructions and video, which may not fully capture all aspects of execution quality. Occlusions and limited temporal resolution can obscure brief harmful events or retries, while final-state recoverability may be difficult to determine from visual evidence alone. Furthermore, subtask progress may not distinguish early actions that do not complete a milestone. Future work could incorporate finer-grained progress annotations, execution time, denser video sampling, and additional camera views or robot-state information to improve evaluation reliability.

\textbf{Context-dependent execution quality.}
REQS uses a fixed scoring rule to support consistent comparisons, but the relative importance of completion, harm, and retries may vary across applications. Specialized or safety-critical deployments may require recalibrating severity definitions and penalty weights against domain-specific consequences. Such adaptations should be validated independently and reported explicitly to distinguish them from the standard score. REQS also assesses how an instruction is executed, rather than whether it is safe or appropriate to execute; task acceptability therefore requires a separate assessment.

\textbf{Downstream generalization and automated assessment.}
Our demonstration-selection experiments are limited to simulated pick-and-place tasks, with REQS computed from simulator state. An important direction is to investigate whether video-derived REQS improves demonstration selection and policy learning on physical robots and more diverse manipulation tasks. The limitations observed in VLM-based assessment also motivate developing more reliable automated judges, particularly for harmful events, retries, and recoverability.

\section*{LLM Usage Statement}
GPT-6.1 Sol was used in the annotation pipeline to generate initial execution labels and subtask checklists and to assess progress from video frames. The resulting annotations were reviewed and refined by the authors.

Large language models were used to assist with editing and refining author-written text and evaluation code.

The authors independently determined the research questions, benchmark design, source inclusion criteria, data curation procedures, metric formulation, experimental design, and interpretation of results. LLMs were not used to generate research hypotheses or conclusions, or to design, conduct, or analyze experiments. The authors take full responsibility for all submitted content.

{
    \small
    \bibliographystyle{ieeenat_fullname}
    \bibliography{main}
}
\input{sec/annotation_appendix}

\end{document}

%% file: preamble.tex
\usepackage[T1]{fontenc}
\usepackage{lmodern}
\usepackage[margin=0.9in]{geometry}
\usepackage{microtype}
\usepackage{graphicx}
\usepackage{amsmath,amssymb}
\usepackage{booktabs,array}
\usepackage[dvipsnames]{xcolor}
\usepackage[numbers,sort&compress]{natbib}
\usepackage[hyphens]{url}
\usepackage[font=small,labelfont=bf,labelsep=period,hypcap=false]{caption}
\usepackage{listings}
\usepackage{enumitem}
\usepackage{titlesec}
\usepackage[breaklinks,colorlinks,allcolors=MidnightBlue]{hyperref}
\usepackage{bookmark}
\hypersetup{
    pdftitle={RoboGrade: Measuring Robotic Execution Quality Beyond Binary Success},
    pdfauthor={Zaruhi Navasardyan, Tatul Danielyan, Mark Hamazaspyan, Hrant Davtyan},
    pdfsubject={},
    pdfkeywords={},
    bookmarksnumbered=true,
    pdfdisplaydoctitle=true
}

\setlist[itemize]{leftmargin=1.5em,itemsep=2pt,topsep=4pt}
\titleformat{\section}{\large\bfseries}{\thesection.}{0.6em}{}
\titleformat{\subsection}{\normalsize\bfseries}{\thesubsection}{0.6em}{}
\titlespacing*{\section}{0pt}{2.2ex plus 1ex minus 0.2ex}{1ex plus 0.2ex}
\titlespacing*{\subsection}{0pt}{1.8ex plus 0.8ex minus 0.2ex}{0.7ex plus 0.2ex}

\makeatletter
\renewcommand{\@maketitle}{%
  \newpage\null\vskip 0.5em
  \begin{center}
    {\fontsize{20}{24}\selectfont\bfseries\@title\par}
    \ifx\@author\@empty\else
      \vskip 1em{\large\@author\par}
    \fi
    \ifx\@date\@empty\else
      \vskip 0.75em{\@date\par}
    \fi
  \end{center}
  \vskip 0.5em
}
\makeatother

%% file: sec/0_abstract.tex
\begin{abstract}
Evaluating robotic manipulation solely through binary task success overlooks substantial differences in execution quality. Executions with identical outcomes may differ in intermediate progress, harmful events, retry attempts, and recoverability. These distinctions are particularly important as vision-language models (VLMs) and learned reward models are increasingly used to evaluate robotic behavior and guide policy improvement. However, existing datasets employ heterogeneous outcome definitions, and it remains unclear whether current evaluation methods adequately capture execution quality beyond task completion. We introduce RoboGrade, a unified framework and benchmark for multidimensional evaluation of robot execution, comprising approximately 2,300 manipulation episodes from 23 sources spanning 15 robotic embodiments and 729 task groups. We establish a task-grounded annotation protocol covering task completion, progress, harmful events and their severity, retries, and recoverability. Building on these dimensions, we propose the Robotic Execution Quality Score (REQS), a standardized 0--5 measure of overall execution quality. We validate REQS against independent human pairwise preferences, achieving 80\% agreement, close to human-to-human consistency as well as validate the downstream utility of REQS for training data selection, especially on small scales. Using this framework, we systematically benchmark open-weight and proprietary VLMs on their ability to assess execution quality from visual evidence. Our results reveal that, despite strong performance in task success recognition, current VLMs struggle to reliably assess broader execution quality dimensions. These findings highlight the gap between recognizing task outcomes and evaluating the quality of robotic behavior. We publicly release the benchmark, annotations, scoring and evaluation framework.
\end{abstract}

%% file: sec/annotation_appendix.tex
\clearpage
\appendix
\pdfbookmark[0]{Appendices}{appendices}
\titleformat{\section}{\large\bfseries}{Appendix~\thesection.}{0.6em}{}

\section{RoboGrade Annotation Protocol}

\subsection{Differences Across Source Annotations}
\begin{table*}[!htbp]
    \centering
    \small
    \caption{Examples of source labels and how the RoboGrade rubric treats the same evidence.}
    \label{tab:source-label-differences}
    \setlength{\tabcolsep}{4pt}
    \begin{tabular}{@{}>{\raggedright\arraybackslash}p{0.19\textwidth}p{0.37\textwidth}p{0.37\textwidth}@{}}
        \toprule
        Source & Source annotation & RoboGrade interpretation \\
        \midrule
        AgiBot World 2026~\cite{agibotworld2026} & An episode can carry an error label for a drop or collision even if the robot subsequently finishes. & Completion is judged against the goal. A corrected drop or collision is recorded separately as harm. \\
        REBOOT~\cite{reboot2026} & Recovered executions are grouped with failures in the source labels used in our collection. & An execution that satisfies the goal is marked complete. Retries and harmful events remain part of its record. \\
        RoboChallenge~\cite{robochallenge2025} & The source provides a completion flag and a separate rubric score that can deduct points for retries. & Completion, progress, and retries are separate labels. A retry affects REQS through its retry term. \\
        ARMBench~\cite{armbench2023} & The labels identify handling defects such as package opening or deconstruction. & These defects provide evidence of severe harm. The completion label independently records whether the object reached the intended destination. \\
        RoboVAD~\cite{robovad2026} & The source identifies anomalies and their types, such as failed grasps or drops. & A repeated grasp is a retry. A drop is labeled for harm, with severity depending on its consequences. Completion is judged separately. \\
        RoboArena~\cite{atreya2025roboarena} & Evaluators mark success and give partial-success scores. A goal can count as reached even if it is later undone. & Completion requires a persistent goal to hold at the final frame. A goal that is reached and then undone counts as progress, and a drop is recorded as harm. \\
        \bottomrule
    \end{tabular}
\end{table*}

Table~\ref{tab:source-label-differences} gives examples of how source labels differ from the RoboGrade rubric. Some sources record different aspects of the same execution, such as an error label for a drop that the robot later corrects. Others define success differently, for example by counting a goal as reached even if it is later undone.

\subsection{Task Grouping and Independent Annotation Passes}

We first partitioned clips by source and task instruction, ignoring capitalization and differences in spacing. Within each partition, we compared grayscale thumbnails of the opening frame from the primary view. Pairwise overlap was the fraction of pixels whose intensity differed by less than 20 on a 0--255 scale. Average-linkage clustering with an overlap threshold of 0.5 separated clips recorded in visually different scenes or camera setups. This procedure uses initial visual similarity as a proxy for a shared recording setup.

GPT-6.1 Sol defined one subtask checklist per group from the instruction and the opening frames of up to four representative clips, using all available views. The model then annotated each execution against the fixed subtask checklist. In a separate pass, it judged task completion directly and annotated execution events. We compared the two passes to flag inconsistencies for human review. For example, a completion judgment and an unmet required goal milestone prompted a full review of the clip, including whether the milestone applied and whether the task criterion was satisfied.

\subsection{Annotation Rounds and Follow-up Evidence}
Table~\ref{tab:annotation-rounds} reports GPT-6.1 Sol annotation rounds across the benchmark. The execution pass covers task completion and the other episode-level labels. The subtask pass assesses individual milestones. A round is a logged model turn, including follow-up attempts that returned no usable answer. For each type of follow-up, we report the number of requests and the number of episodes with at least one such request. A single episode can contribute multiple requests of either type.

\begin{table*}[!htbp]
    \centering
    \small
    \caption{Annotation rounds and follow-up requests by GPT-6.1 Sol. Round columns count episodes. The last two columns show requests (distinct episodes).}
    \label{tab:annotation-rounds}
    \begin{tabular}{llrrrrr}
        \toprule
        Subset & Annotation pass & 1 round & 2 rounds & 3 rounds & Zoom & Additional frames \\
        \midrule
        Benchmark (2,301)     & Execution & 805 & 969 & 527 & 1,085 (692) & 3,292 (1,375) \\
                           & Subtasks & 2,053 & 195 & 53 & 294 (178) & 188 (137) \\
        \bottomrule
    \end{tabular}
\end{table*}

\subsection{Label Definitions}

\paragraph{Task completion.}
An execution is complete when it meets the observable success criterion derived from the task instruction. For goals that require a persistent change, the required state must hold at the end of the clip. A momentary goal such as pressing a button is assessed when it occurs. Harm and retries are recorded separately, so an execution may complete the task after a drop, spill, or repeated attempt.

\paragraph{Progress.}
All attempts in a task group share a subtask checklist of observable milestones. A control milestone is achieved when the robot gains control of a relevant object, for example by lifting it or moving it from its initial position. It counts even if the robot later loses control of the object. A goal milestone specifies a task-relevant state, such as an object remaining in its destination after release. Persistent goals must hold at the final frame. Progress is the fraction of applicable items achieved.

\paragraph{Harm.}
Each observable harmful event is recorded with its onset, type, and severity. The event types include premature drops, unintended collisions, objects knocked over or displaced, spills, damage, objects leaving the workspace, contact initiated by the robot with a person, and the robot becoming stuck. Minor harm includes events such as a bump, an intact drop within reach, or a small localized spill. Severe harm includes damage, substantial spills, loss of access to an object, contact initiated by the robot with a person, or immobilization after a collision. REQS uses the highest severity harm that occurred during the execution for scoring. 

\paragraph{Retries.}
A retry begins when the robot makes a distinct new attempt at a step it has already tried, such as returning to regrasp an object or attempting an insertion again. Continuous adjustments within one attempt are treated as part of that attempt. Retries are timestamped at the start of the renewed attempt. The episode-level count is recorded as zero, one, two, or three or more.

\paragraph{Recoverability.}
For an incomplete execution, the final scene is assessed for whether the robot could still finish without a person changing the scene. An intact object within reach may allow recovery. A required object that is damaged or out of reach, or a robot immobilized by a collision, may prevent it. This label concerns the feasibility of continuing from the observed state. It is inapplicable to completed executions.

\subsection{Agreement in Human Review}

We conducted a human review of 1156 episodes, with reviewers having access to the initial annotations. Of these, 756 episodes were flagged for review based on the three criteria described in Section~3.3 of the main paper. Agreement between the initial and final labels for these flagged episodes was 73.2\% for task completion, 82.4\% for harm severity, 89.4\% for retry category, and 90.8\% for recoverability among failed executions.

To assess whether the review criteria overlooked annotation errors, we additionally audited 400 randomly selected episodes from the unflagged pool, which had not triggered any of the review criteria. In this audit, human reviewers accepted 90\% of episodes without any label modifications, indicating high agreement on unflagged episodes.

\begingroup
    \section{REQS: Worked Examples and Human Validation}
    \label{sec:supp-worked-reqs}
    {\centering\includegraphics[width=0.92\textwidth]{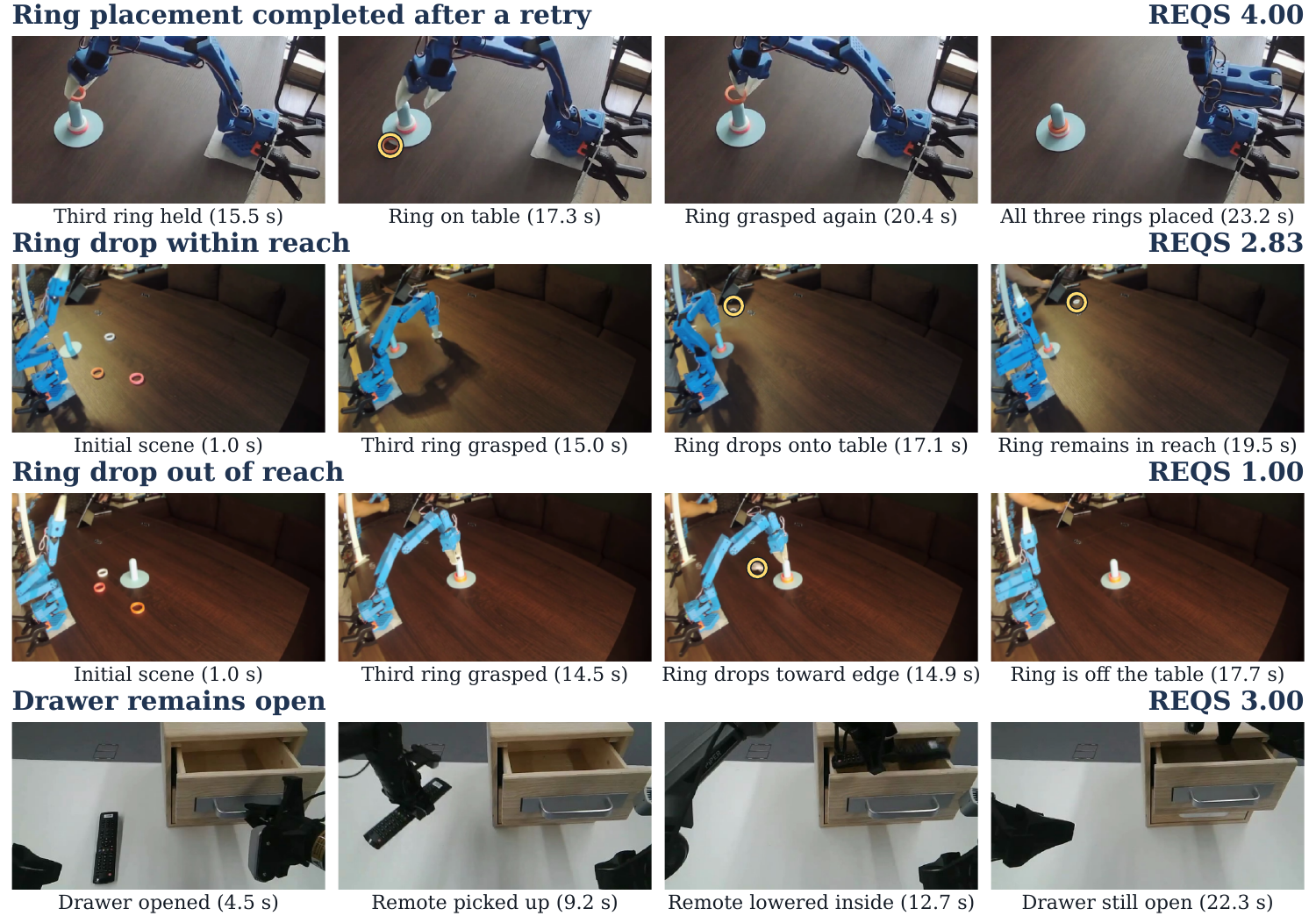}\par}
    \captionof{figure}{REQS for four executions from RoboVAD~\cite{robovad2026} and ViFailback~\cite{vifailback2025}. Row 1 shows completion after a retry. Rows 2 and 3 show ring drops that leave the task recoverable and unrecoverable, respectively. Row 4 shows a recoverable failure in a drawer task. Circles mark dropped rings.}
    \label{fig:supp-execution-examples}
    \vspace{0.6em}
\endgroup

\paragraph{Completion with a retry and minor harm.}
In Figure~\ref{fig:supp-execution-examples}, row 1, dropping the third ring onto the table counts as minor harm because it stays within reach. Regrasping it counts as one retry. Placing all three rings gives a base score of 5. The drop and retry each cost 0.5, yielding $5-0.5-0.5=4$.

\paragraph{Partial progress after a ring drop.}
In Figure~\ref{fig:supp-execution-examples}, rows 2 and 3, each ring has a pickup and a placement subtask. Both executions complete three pickups and two placements, giving $5/6$ progress, and neither has a retry. For incomplete tasks, the base score is 4 times progress. In row 2, the ring lands within reach, so the task remains recoverable and the drop costs 0.5 for minor harm. REQS is $4\times(5/6)-0.5\approx2.83$. In row 3, the ring falls off the table and out of reach, incurring deductions of 2 for severe harm and 1 for an unrecoverable final state. The score before applying the minimum is $4\times(5/6)-2-1\approx0.33$. Since the robot made progress, the one-point minimum gives a REQS of 1.

\paragraph{Partial progress with an open drawer.}
In Figure~\ref{fig:supp-execution-examples}, row 4, the robot opens the drawer, picks up the remote, and places it inside, completing three of four subtasks. The drawer remains open and can still be closed. With no harm or retries, REQS is $4\times(3/4)=3$.

\subsection{Human Preference Validation}

The REQS penalty weights were calibrated by the authors using a separate set of executions. Human preference validation was conducted by two independent reviewers who were not involved in the calibration process. Reviewers received only the task instruction and execution videos, without access to REQS scores or component annotations.

\section{Analysis of Annotation Dimensions}
\label{sec:supp-vlm-results}

We select the five judges with the lowest REQS MAE in Section~5.2: Claude Opus 5.5, Qwen3.8-27B, Gemini 3.8 Flash, TypeSafe Jev 1.14, and GLM 5.3 Flash. We apply the component scoring rules in Section~5.1 to the same 2,301 clips. Missing or uncertain judgments count as incorrect.

\paragraph{Completion and subtasks.}
Claude leads on balanced accuracy by recognizing more successes and achieved subtasks. Gemini is better at flagging failed executions (87.8\% recall vs.\ 84.0\%) and unachieved subtasks (80.6\% vs.\ 77.5\%). Jev shows the widest gap between subtask classes, recognizing 93.2\% of achieved subtasks and 58.1\% of unachieved ones. The tendency to overstate progress is especially visible on ARMBench. All five judges call 21 of 33 failed executions complete and mark 40 of 53 unachieved subtasks as achieved, even though GLM alone catches three reference failures the other four miss.

\paragraph{Harm and retries.}
The five judges share errors on 63 of 132 severe-harm clips and 141 of 279 clips with three or more retries. GLM is the only judge to correctly label 33 minor-harm clips, but resolves only one additional severe-harm case that the other four miss. The relative strengths of judges also vary by source. Qwen labels 16 of 24 severe-harm clips on RoboVAD, compared with 12 for Claude, but labels one of 48 on ARMBench, compared with nine for Claude. Jev identifies the highest retry category on 63 of 64 PhAIL clips and on 4 of 215 such clips elsewhere.

At the event level, similar recall can come with very different numbers of unmatched predictions. Claude and Jev match similar shares of reference retries (49.2\% and 47.1\%), while their precision differs substantially (68.9\% and 27.4\%, Table~\ref{tab:supp-event-localization}). The harm results illustrate the same trade-off. GLM matches 40.2\% of reference harm events compared with Gemini's 36.0\%, while 41.1\% of GLM's predicted events match the reference, compared with 75.8\% for Gemini.

\begin{table}[!htbp]
    \centering
    \small
    \caption{Event localization precision and recall (P / R) in percent. These measures compare individual events on clips with usable event lists.}
    \label{tab:supp-event-localization}
    \begin{tabular}{@{}lrr@{}}
        \toprule
        Judge & Harm P / R & Retry P / R \\
        \midrule
        Claude Opus 5.5 & 69.3 / 47.6 & 68.9 / 49.2 \\
        Gemini 3.8 Flash & 75.8 / 36.0 & 63.0 / 33.1 \\
        Qwen3.8-27B & 67.5 / 36.6 & 59.9 / 35.4 \\
        TypeSafe Jev 1.14 & 41.7 / 39.5 & 27.4 / 47.1 \\
        GLM 5.3 Flash & 41.1 / 40.2 & 47.0 / 29.0 \\
        \bottomrule
    \end{tabular}
\end{table}

\paragraph{REQS error by source.}
Table~\ref{tab:vlm-dataset-mae} shows how aggregate score errors vary across sources. On ARMBench, where the judges often miss incomplete executions, every judge has an MAE above one REQS point. On REBOOT, Claude's MAE is 0.05.

\paragraph{Agreement and voting.}
We retrospectively test plurality voting for each component on the saved annotations. Missing, uncertain, and inapplicable labels abstain. We break ties in the order Claude, Gemini, Qwen, Jev, GLM, then evaluate the selected label and its associated events under the original rules. The five-judge vote improves recovery balanced accuracy over Claude alone (72.9\% vs.\ 65.3\%) and yields similar subtask accuracy (86.0\% vs.\ 85.8\%). Its completion, harm, and retry scores are lower.

\begingroup
    \centering
    \small
    \setlength{\tabcolsep}{1.5pt}
    \captionof{table}{Mean absolute error in REQS points by source dataset. Clips is the number of reference-scored clips. Bold indicates the lowest error in each row before rounding. Open denotes open-weight models, and Closed denotes proprietary models. Parameter counts are totals. Nemotron activates 12B of its 120B parameters, and GLM 5.3 Flash activates 18B of its 320B parameters.}
    \label{tab:vlm-dataset-mae}
    \resizebox{\textwidth}{!}{%
    \begin{tabular}{lrrrrrrrrrrrr}
        \toprule
        Dataset & Clips & \shortstack{Claude\\Opus 5.5\\\scriptsize Closed} & \shortstack{Gemini\\3.8 Flash\\\scriptsize Closed} & \shortstack{Qwen3.8\\27B\\\scriptsize Open} & \shortstack{Jev\\1.14\\\scriptsize Closed} & \shortstack{GLM 5.3\\Flash\\320B\\\scriptsize Open} & \shortstack{Qwen3.5\\9B\\\scriptsize Open} & \shortstack{Gemma 4\\31B\\\scriptsize Open} & \shortstack{Qwen3.5\\4B\\\scriptsize Open} & \shortstack{Nemotron\\3.5 Super VL\\120B\\\scriptsize Open} & \shortstack{Cosmos-\\Reason2\\32B\\\scriptsize Open} & \shortstack{Gemma 4\\12B\\\scriptsize Open} \\
        \midrule
        HRDexDB & 179 & \textbf{0.88} & 0.95 & 1.05 & 1.44 & 0.94 & 1.13 & 1.09 & 1.21 & 1.44 & 1.48 & 1.73 \\
        RoboVAD & 176 & 0.21 & 0.33 & \textbf{0.15} & 0.40 & 0.32 & 0.41 & 0.65 & 0.45 & 0.82 & 0.89 & 1.31 \\
        AgiBot World 2026 & 167 & 0.54 & 0.63 & 0.60 & \textbf{0.38} & 0.50 & 0.62 & 1.14 & 0.77 & 0.75 & 0.85 & 0.84 \\
        ArmnetBench & 157 & 0.26 & 0.33 & 0.52 & \textbf{0.23} & 0.74 & 0.95 & 1.13 & 1.04 & 1.60 & 1.88 & 2.13 \\
        RoboChallenge & 153 & \textbf{0.18} & 0.26 & 0.41 & 0.30 & 0.61 & 0.79 & 0.62 & 0.76 & 1.19 & 1.44 & 1.81 \\
        ARMBench & 139 & \textbf{1.24} & 1.53 & 1.44 & 1.31 & 1.56 & 1.39 & 1.50 & 1.50 & 1.54 & 1.45 & 1.45 \\
        RH20T & 127 & \textbf{0.49} & 0.82 & 0.82 & 0.81 & 0.81 & 0.95 & 0.95 & 0.91 & 1.68 & 1.66 & 1.90 \\
        REBOOT & 117 & \textbf{0.05} & 0.08 & 0.15 & 0.38 & 0.55 & 0.62 & 1.66 & 0.45 & 0.32 & 0.78 & 0.38 \\
        REASSEMBLE & 113 & \textbf{1.21} & 1.85 & 1.80 & 2.05 & 2.13 & 1.99 & 2.17 & 1.75 & 1.84 & 2.16 & 2.66 \\
        BotFails & 108 & \textbf{0.55} & 0.63 & 0.80 & 0.60 & 0.86 & 1.08 & 0.89 & 0.98 & 1.29 & 1.71 & 1.70 \\
        Other & 102 & \textbf{0.25} & 0.63 & 0.73 & 0.56 & 1.00 & 0.69 & 0.98 & 1.08 & 2.08 & 1.70 & 2.14 \\
        FailTime-Long & 92 & \textbf{0.40} & 0.40 & 0.58 & 0.78 & 0.81 & 0.98 & 1.07 & 1.07 & 1.44 & 1.88 & 1.89 \\
        Robometer & 92 & \textbf{0.23} & 1.22 & 0.76 & 0.61 & 1.44 & 1.51 & 1.23 & 1.36 & 1.86 & 2.35 & 2.34 \\
        RoboReward & 78 & \textbf{0.16} & 0.96 & 0.67 & 0.54 & 1.05 & 0.92 & 1.00 & 1.02 & 1.52 & 1.50 & 1.56 \\
        ViFailback & 77 & \textbf{0.16} & 0.21 & 0.19 & 0.23 & 0.19 & 0.37 & 0.73 & 0.45 & 0.90 & 1.15 & 1.09 \\
        TOTO & 76 & 0.82 & 1.68 & 0.84 & \textbf{0.82} & 1.59 & 1.39 & 1.30 & 1.54 & 1.83 & 1.52 & 1.72 \\
        RoboArena & 73 & \textbf{0.49} & 0.62 & 0.70 & 1.11 & 1.30 & 1.36 & 1.19 & 1.50 & 2.34 & 2.78 & 2.89 \\
        HapticWAM & 69 & \textbf{0.29} & 0.35 & 0.80 & 0.68 & 0.73 & 0.70 & 0.70 & 0.88 & 1.98 & 1.29 & 2.26 \\
        PhAIL & 69 & \textbf{0.30} & 0.38 & 0.41 & 0.45 & 0.54 & 0.93 & 0.40 & 1.14 & 0.80 & 2.94 & 1.14 \\
        RoboFAC & 57 & 0.20 & \textbf{0.16} & 0.30 & 0.38 & 0.25 & 0.43 & 0.50 & 0.60 & 1.48 & 1.43 & 1.62 \\
        Handover Failure Detection & 46 & \textbf{0.39} & 0.64 & 0.47 & 1.01 & 0.67 & 1.27 & 1.12 & 1.20 & 2.10 & 1.73 & 2.07 \\
        RoboFail (REFLECT) & 20 & 1.23 & 1.29 & 1.86 & \textbf{1.17} & 1.41 & 1.95 & 1.65 & 1.70 & 1.95 & 2.26 & 2.15 \\
        Humanoid competition & 14 & \textbf{0.58} & 0.61 & 0.62 & 1.79 & 2.17 & 1.81 & 1.96 & \textbf{0.58} & 2.00 & 2.07 & 0.67 \\
        \bottomrule
    \end{tabular}%
    }\par
    \vspace{0.6em}
\endgroup

\section{VLM Evaluation Details}

\subsection{Visual Inputs and Inference Settings}
The benchmark contains 4,640 camera-view videos across 2,301 episodes. Each model receives all available camera views for each clip. For models that sample at a fixed frame rate, the number of input frames depends on the duration of each view.

\paragraph{Claude Opus 5.5.}
We sample 32 evenly spaced JPEG frames per view and limit the long image edge to 640 pixels. Each frame is preceded by its \verb|mm:ss| timestamp. We run Claude Opus 5.5 through an API. The default prompt precedes the labeled views.

\paragraph{Gemini 3.8 Flash.}
We send one labeled video per view after the default prompt. The API samples each video at one frame per second at its default resolution and provides timestamps. We use medium thinking and seed 0.

\paragraph{Qwen3.8-27B.}
We send one labeled video per view, sampled at two frames per second, after the prompt. We replace the \verb|mm:ss| references in the input description and both event timing questions with timestamps in seconds (for example, \verb|<1.3 seconds>|). We enable thinking and use temperature 1.0, top-$p$ of 0.95, top-$k$ of 20, and seed 0.

\paragraph{Qwen3.5-9B and Qwen3.5-4B.}
We use the same prompt variant and two-frame-per-second sampling as Qwen3.8-27B. We enable thinking with temperature 1.0, top-$p$ of 0.95, top-$k$ of 20, presence penalty 1.5, and seed 0. The output limit is 32,768 tokens.

\paragraph{Gemma 4 31B.}
We send one labeled video per view, sampled to 32 frames, after the default prompt. We enable thinking and use temperature 0.

\paragraph{Gemma 4 12B.}
We use the same prompt, 32 frames per view, thinking mode, and temperature 0 as Gemma 4 31B. The output limit is 32,768 tokens.

\paragraph{TypeSafe Jev 1.14.}
We use the same 32 timestamped frames per view as Claude and send one request per clip in long-context mode. We translate the rubric into typed questions about completion, each subtask, recoverability, harm and retry counts, and events in each interval between frames. For each interval we ask for the harm type and severity and whether a retry began and for which item. We threshold binary answers at probability 0.5 and mark all items achieved when the model predicts completion. For harm and retries, we take the most probable count and select that many intervals with the highest event probabilities. We assign each selected event the timestamp of the later frame.

\paragraph{GLM 5.3 Flash.}
 We send one labeled video per view after the default prompt, through OpenRouter, pinned to Z.AI's own endpoint. We set no sampling parameters, so Z.AI's defaults apply (it recommends temperature 1 and top-p 0.95), with thinking always on at maximum reasoning effort, up to 131,072 output tokens and no seed.

\paragraph{Nemotron 3.5 Super VL.}
We use the 120B-parameter checkpoint with 12B active parameters. We send one video per camera view at two frames per second with a cap of 256 frames per video and a pruning rate of 0.5. We adapt the Qwen prompt variant to the processor's timestamp format (\texttt{Frame N sampled at X.XX seconds}). We enable reasoning and use temperature 0.6, top-$p$ of 0.95, seed 0, and an output limit of 32,768 tokens. We run vLLM in BF16 on two B200 GPUs with tensor parallelism of two and five speculative draft tokens.

\paragraph{Cosmos-Reason2-32B.}
We sample each video at four frames per second. We send the labeled views before the Qwen timestamp variant of the prompt and set the system message to \texttt{You are a helpful assistant.} We append an instruction to enclose the reasoning in \verb|<think>| and \verb|</think>| tags and place the final JSON answer immediately after the closing tag. We use temperature 0.6, top-$p$ of 0.95, top-$k$ of 20, and seed 0. The generation limit, is 4,096 tokens.

\clearpage
\subsection{Default Judge Prompt}
We insert the task instruction and numbered subtask checklist into the template below. The double braces escape literal JSON braces in the source template.

\begin{lstlisting}[
    basicstyle=\fontsize{9}{11}\selectfont\ttfamily,
    breaklines=true,
    breakindent=0.75em,
    columns=fullflexible,
    keepspaces=true,
    frame=single,
    xleftmargin=0.75em,
    xrightmargin=0.75em,
    aboveskip=0.25em,
    belowskip=0pt
]
You judge one robot video. Answer only from the video.

INPUT
For each camera view: the video, given as frames in time order, each preceded by its time in the clip (mm:ss). Each view is named by its camera where known: a wrist camera is mounted on the arm and moves with the gripper; a fixed or head camera shows the scene from outside the gripper.

TASK
Instruction: {instruction}
Subtasks:
{subtasks}

QUESTIONS
1. done: is the instruction completed at the end of the clip? Problems along the way do not change this.
2. subtasks: for each checklist item, one of:
   "done": the subtasks was done anytime in the clip.
   "not_done": the task condition is not met.
   A subtask that removes, takes out, opens, closes or otherwise changes an object is done if that state holds. If the instruction names the arm or hand for a step, only that arm or hand counts. If the instruciton is to put, release, place, move etc the girpper should release the object to have successfully done the task.
3. harm_events: every harm event, each listed separately, in time order. For each: "time_s", when the event starts (the object begins to leave the gripper, the first contact, the object begins to tip or move, contents first leave their container, the object leaves the table or the view, the robot first stops moving), in seconds, read from the mm:ss before the first frame that shows.
Harm events (type):
   - drop: an object leaves the gripper before it is placed, whether or not it lands intact.
   - collision: the robot, or an object it holds, hits something other than its target. Record it even when nothing breaks or moves.
   - knocked_over: an object that was standing falls on its side or tips over.
   - displaced: an object the robot did not mean to move is pushed or dragged out of its place (a container pushed across the table, a cube nudged along it).
   - spill: contents (liquid, powder, grains, small items) leave their container.
   - damage: an object is broken, torn, deformed, crushed, or its packaging or cover opens (a box flap or a book cover that came open).
   - object_off_workspace: an object falls off the table, shelf, bin or conveyor, or leaves the camera view and is not seen again by the end of the clip.
   - person_contact: the robot, or an object it holds, hits a person. A person touching or handing over an object, or touching the robot, is not harm.
   - robot_stuck: the robot is jammed or caught after a collision and cannot move on.
   - other.
   Severity of each event:
   - minor: no lasting effect: the objects are intact, usable and still in the robot's reach at the end (an intact drop, a bump, an object knocked over or pushed but undamaged), or a small spill: a few drops or a small amount that stays in one small patch on the table.
   - severe: a lasting effect that the robot cannot undo: a large spill (it spreads over the table, reaches the floor, a person or a device, or empties a large part of the container), damage, an object off the workspace or out of view for good, contact with a person, or the robot stuck.
4. retry_events: every retry, each listed separately, in time order. A retry is a new attempt at a step it already tried: it re-grasps, approaches the same object or part again, or repeats a turn, push, insertion or placement that did not take effect. A retry does not need a visible drop or missed grasp: if the gripper leaves an object or part it was working on and comes back to do the same thing again, it is a retry. It is a retry whether or not a person helped in between (picked the object up, set it upright, moved it into reach). Not retries: small corrections inside one continuous motion (slowing down, adjusting the approach while still moving toward the object). For each: "time_s", when the new attempt starts (the robot begins to go back to the object or part it already tried), in seconds, read from the mm:ss before the first frame that shows it; "subtask", the number of the subtask in the list above that the robot tries again.
5. recoverable: only if done is "no"; judged at the last frame. "yes" if the robot could still finish the task alone, without a person changing the scene: the objects it needs are intact and inside its reach and the robot can still move. "no" if the visible state makes the task impossible for the robot alone: an object off the table or out of reach, an object broken, contents spilled, liquid poured in the wrong place, the robot stuck after a collision. Judge the state, not what the robot does next. "not_applicable" if done is "yes".

Answer "cant_tell" for any question the frames do not settle.

OUTPUT
One JSON object and nothing else:
{{"done": "yes|no|cant_tell", "subtasks": ["done|not_done|cant_tell", ...], "harm_events": [{{"time_s": 0.0, "type": "drop|collision|knocked_over|displaced|spill|damage|object_off_workspace|person_contact|robot_stuck|other", "severity": "minor|severe", "what": "..."}}, ...], "retry_events": [{{"time_s": 0.0, "subtask": 1}}, ...], "recoverable": "yes|no|not_applicable|cant_tell"}}
"subtasks" has one entry per checklist item, in checklist order. "harm_events" and "retry_events" are lists ([] if none), or "cant_tell" if the frames do not settle whether any happened. "recoverable" is "not_applicable" when done is "yes".
\end{lstlisting}

\begingroup
\clearpage
\section{Benchmark Sources}
Table~\ref{tab:benchmark-sources} lists the 23 sources represented in RoboGrade. Counts refer to episodes in the frozen benchmark. Success and failure use the RoboGrade completion labels. The source descriptions identify the origin of the clips and the types of executions included.

\begingroup
    \centering
    \small
    \captionof{table}{Sources of the 2,301 real-world episodes in RoboGrade.}
    \label{tab:benchmark-sources}
    \setlength{\tabcolsep}{4pt}
    \begin{tabular}{@{}lrrrp{0.46\textwidth}@{}}
        \toprule
        Source & Success & Failure & Total & Content in RoboGrade \\
        \midrule
        AgiBot World 2026~\cite{agibotworld2026} & 150 & 17 & 167 & Manipulation in household and work settings. \\
        ARMBench~\cite{armbench2023} & 106 & 33 & 139 & Bin-to-bin transfers with package defect annotations. \\
        ArmnetBench~\cite{armnetbench2026} & 83 & 74 & 157 & Manipulation rollouts with operator quality ratings. \\
        BotFails~\cite{botfails2026} & 31 & 77 & 108 & Teleoperated tasks with staged manipulation anomalies. \\
        FailTime-Long~\cite{failtimedata} & 40 & 52 & 92 & Extended manipulation clips with temporal failure annotations. \\
        HapticWAM~\cite{hapticwam2026} & 32 & 37 & 69 & Pick-and-place trials, including fragile-object handling. \\
        Handover Failure Detection~\cite{hfd2024} & 25 & 21 & 46 & Robot-to-human and human-to-robot handovers. \\
        HRDexDB~\cite{hrdexdb2026} & 102 & 77 & 179 & Physical robot grasps with dexterous hands. \\
        Humanoid competition & 8 & 6 & 14 & Selected clips from humanoid competition recordings. \\
        Other & 49 & 53 & 102 & Selected real-robot clips from research project pages. \\
        PhAIL~\cite{phail2026} & 13 & 56 & 69 & Multi-object transfers with item-level completion counts. \\
        REASSEMBLE~\cite{reassemble2025} & 56 & 57 & 113 & Contact-rich assembly and disassembly segments. \\
        REBOOT~\cite{reboot2026} & 114 & 3 & 117 & Precision assembly, including recovered executions. \\
        RoboFail (REFLECT)~\cite{reflect2023} & 5 & 15 & 20 & Real-robot manipulation failures and explanations. \\
        RH20T~\cite{rh20t2024} & 73 & 54 & 127 & Multiview recordings of everyday manipulation skills. \\
        RoboArena~\cite{atreya2025roboarena} & 25 & 48 & 73 & Paired real-robot policy rollouts with human preferences. \\
        RoboChallenge~\cite{robochallenge2025} & 81 & 72 & 153 & Real-robot evaluations across standardized tasks. \\
        RoboFAC~\cite{robofac2025} & 31 & 26 & 57 & Real-robot episodes with failure diagnosis annotations. \\
        Robometer~\cite{liang2026robometer} & 38 & 54 & 92 & Real-robot rollouts drawn from a reward-learning corpus. \\
        RoboReward~\cite{lee2026roboreward} & 41 & 37 & 78 & Robot videos with task-progress scores. \\
        RoboVAD~\cite{robovad2026} & 146 & 30 & 176 & Robot manipulation videos with anomaly annotations. \\
        TOTO~\cite{toto2023} & 37 & 39 & 76 & Scooping trials with outcome measurements. \\
        ViFailback~\cite{vifailback2025} & 47 & 30 & 77 & Bimanual manipulation clips with failure diagnoses. \\
        \midrule
        Total & 1,333 & 968 & 2,301 & \\
        \bottomrule
    \end{tabular}
\par\endgroup
\vspace{0.6em}
\endgroup

\clearpage
\section{Downstream Validation Details}
\label{sec:supp-downstream}
\subsection{Physical Load of the Trained Policies}
Table~\ref{tab:policy-load} reports the physical load on the robot for the BC-ResNet policies of Section 4.3, on the same 200 test scenes as Table 2.
\begin{itemize}
    \item Gripper closures count how often the policy commands the gripper to close.
    \item Contact impulse is the contact force between the robot and the table or non-task objects, summed over time across the episode.
    \item Duration is the length of the episode.
    \item Hand path is the distance traveled by the gripper.
\end{itemize}

\noindent\begin{minipage}{\columnwidth}
    \centering
    \small
    \captionof{table}{Physical load of BC-ResNet test rollouts by training-data selection strategy. Values are means per test episode. Rows are the training-set sizes of Table 2, given as the number of time steps (frames).}
    \label{tab:policy-load}
    \setlength{\tabcolsep}{3pt}
    \begin{tabular}{@{}llrrrr@{}}
        \toprule
        Time & & Gripper & Contact & & Hand \\
        steps & Strategy & closures & impulse & Duration & path \\
        & & & (N\,s) & (s) & (m) \\
        \midrule
        12k & Random & 5.07 & 17.4 & 11.24 & 1.34 \\
         & Success only & 3.30 & 13.4 & 10.08 & 1.31 \\
         & Top REQS & 2.03 & 7.2 & 9.28 & 1.25 \\
        \midrule
        24k & Random & 2.91 & 9.0 & 10.04 & 1.29 \\
         & Success only & 2.01 & 8.2 & 9.45 & 1.28 \\
         & Top REQS & 1.56 & 6.1 & 9.05 & 1.25 \\
        \midrule
        47k & Random & 1.51 & 5.5 & 8.88 & 1.24 \\
         & Success only & 1.56 & 6.0 & 8.91 & 1.25 \\
         & Top REQS & 1.35 & 6.1 & 8.65 & 1.23 \\
        \bottomrule
    \end{tabular}
\end{minipage}\par

With 12k training time steps, top-REQS policies closed the gripper 38\% less often than success-only policies, had 46\% less contact impulse, and finished 8\% faster. The differences became smaller as the training set grew (Table~\ref{tab:policy-load}).

The lower load mostly reflects fewer failed and low-scoring episodes, which account for most of the contact and gripper use. Episodes with REQS 5 had almost no contact with the table or other objects.